\documentclass[a4paper, fleqn]{cas-sc}
\usepackage{multirow}

\usepackage[dvipsnames]{xcolor}
\usepackage{comment}
\usepackage{caption}
\usepackage{ragged2e}
\usepackage{graphicx}
\usepackage{verbatim}
\usepackage{amssymb}
\usepackage{float}
\usepackage{makecell}
\usepackage{setspace} 
\let\printorcid\relax

\usepackage[numbers]{natbib}

\def\tsc#1{\csdef{#1}{\textsc{\lowercase{#1}}\xspace}}
\tsc{WGM}
\tsc{QE}

\shorttitle{From Zero to Hero: Harnessing Transformers for Biomedical
Named Entity Recognition in Zero- and Few-shot Contexts}    

\shortauthors{Košprdić et al.}  

\title[mode = title]{From Zero to Hero: Harnessing Transformers for Biomedical Named Entity Recognition in Zero- and Few-shot Contexts}  

\author[inst1]{Miloš Košprdić}[orcid=0000-0001-6902-3639]
\author[inst1]{Nikola Prodanović}[orcid=0000-0001-8698-9830]
\author[inst1]{Adela Ljajić}[orcid=0000-0001-7326-059X]
\author[inst1]{Bojana Bašaragin}[orcid=0000-0002-7679-1676]
\author[inst1,inst2]{Nikola Milošević\corref{cor1}}[orcid=0000-0003-2706-9676]\cormark[*]\ead{nikola.milosevic@bayer.com}

\affiliation[inst1]{organization={Institute for Artificial Intelligence Research and Development of Serbia},
            addressline={Fruškogorska 1}, 
            city={Novi Sad},
            postcode={21000}, 
            country={Serbia}}
            
\affiliation[inst2]{organization={Bayer A.G., Reaserch and Development},
            addressline={Mullerstrasse 173}, 
            city={Berlin},
            postcode={13342},
            country={Germany}}

\cortext[cor1]{Corresponding author, contact email: nikola.milosevic@bayer.com}

\begin{abstract}
Supervised named entity recognition (NER) in the biomedical domain depends on large sets of annotated texts with the given named entities. The creation of such datasets can be time-consuming and expensive, while extraction of new entities requires additional annotation tasks and retraining the model. This paper proposes a method for zero- and few-shot NER in the biomedical domain to address these challenges. The method is based on transforming the task of multi-class token classification into binary token classification and pre-training on a large number of datasets and biomedical entities, which allows the model to learn semantic relations between the given and potentially novel named entity labels. We have achieved average F1 scores of 35.44\% for zero-shot NER, 50.10\% for one-shot NER, 69.94\% for 10-shot NER, and 79.51\% for 100-shot NER on 9 diverse evaluated biomedical entities with fine-tuned PubMedBERT-based model. The results demonstrate the effectiveness of the proposed method for recognizing new biomedical entities with no or limited number of examples, outperforming previous transformer-based methods, and being comparable to GPT3-based models using models with over 1000 times fewer parameters. We make models and developed code publicly available. 
\bigbreak

\end{abstract}

\begin{document}



\maketitle
\section{Introduction}
\label{sec:intro}

Supervised machine learning algorithms rely heavily on large volumes of annotated data, which can be difficult and costly to obtain due to the manual effort and expertise required for accurate labeling\citep{van2020survey}. This difficulty is particularly evident when dealing with textual data in the biomedical field. Biomedical named entity recognition (NER) plays a crucial role in disease comprehension, target, and drug discovery, as well as precision medicine. Annotating novel entities, for which there is no previous data, is often necessary.

Recognizing named entities, such as gene names, drugs, chemicals, diseases, biomarkers, cells, cell lines, tissues, organs, DNA or RNA sequences, holds great value in various applications. These include information retrieval \citep{lu2009evaluation}, de-identification of medical records \citep{milosevic2020mask,dehghan2015combining,kovavcevic2024identification}, relationship extraction \citep{milovsevic2023comparison}, knowledge graph creation \citep{luo2022biored}, question answering \citep{toral2005improving}, automatic summarization \citep{aramaki2009text2table}, and many others.

In the biomedical domain, the costs associated with text annotation are especially high due to the need for professional annotators such as medical doctors, biologists, and chemists. Estimations suggest that annotating a single instance can range from \$0.71 to \$377, depending on the complexity and number of annotators involved \cite{carrell2016juice}. In our previous work \citep{milovsevic2023comparison}, we paid 2.6 euros per annotated sentence. Considering the substantial number of annotated examples required for each entity type to train a reliable machine learning model for NER, these costs can escalate rapidly.

Zero-shot and few-shot machine learning algorithms can help address issues with a lack of annotated data. Zero-shot learning for NER allows recognizing named entities in text based on no given examples during the training \citep{xian2018zero, nguyen2021dozen}, while few-shot learning allows training a more efficient model for NER, based on a few annotated example sentences \citep{hofer2018few, fritzler2019few,moscato2023multi}. These types of machine learning leverage transfer from other domains and previous pre-training to build more efficient models. Several zero-shot and few-shot algorithms have been proposed in the last several years for medical coding \citep{ziletti2022medical}, general NER \citep{aly2021leveraging}, and mix of domains, such as literature, music, politics, and natural science \cite{nguyen2021dozen}. However, the area of zero-shot and few-shot machine learning for NER remains underexplored, particularly within the biomedical domain. This domain presents unique challenges due to its specialized vocabulary, which includes Latin and Greek terminology, as well as the heavy use of acronyms and abbreviations. Additionally, the diversity and complexity of biomedical entities surpass those found in general contexts, necessitating tailored approaches and specialized language models for effective entity recognition.

In this paper, we present an approach for biomedical NER based on transforming the task of multi-class token classification into binary token classification and pre-training on a large number of datasets and biomedical entities. The pre-trained initial model for zero-shot learning is based on transformer architecture and utilizes domain-specific BERT-based language models (BioBERT \citep{Lee2020} and PubMedBERT \citep{gu2021domain}).

\section{Background}
\label{sec:bg}

Named entity recognition (NER) is an information extraction task focused on identifying mentions of specific information units, or named entities, in text. While there is no consensus on the exact definition of named entity \citep{Marrero2013}, it generally includes generic units like personal names, locations, and organizations, as well as domain-specific entities such as gene, protein, or enzyme names \citep{Li2020}. Early NER research from the 1990s primarily addressed identifying personal, location, and organization names in news reports, scientific texts, and later, emails \citep{Nadeau2007}. Initial systems relied on handcrafted rules, such as regular expressions \citep{appelt1993fastus} or syntactic-lexical patterns \citep{morgan1995university, grishman1995nyu}, often supplemented with gazetteers \citep{iwanska1995wayne}. Although these systems achieved high precision, they were limited by the quality and scope of the rules and required significant time and expertise to develop. As a result, subsequent approaches shifted toward machine learning methods, using fully annotated texts (supervised) \citep{zhou2002named, curran2003language, mccallum2003early, li2005svm}, small sets of entities and contexts (semi-supervised) \citep{carreras2002named, agerri2016robust}, or context and lexical resources alone (unsupervised) \citep{etzioni2005unsupervised, munro2012accurate}. Today, state-of-the-art NER performance is achieved using deep learning techniques \citep{yadav2019survey, Li2020}.
    
The early 2000s saw the release of the first publicly available biomedical corpora \citep{kim2003genia, hirschman2005overview}, which accelerated research on recognizing entities like proteins, genes, and drugs in biomedical texts, often driven by shared tasks such as BioCreative \citep{hirschman2005overview} and JNLPBA \citep{collier2004introduction}. Early biomedical NER systems included rule-based approaches \citep{gaizauskas2003protein} and various machine learning methods like Naive Bayes \citep{nobata1999automatic}, Support Vector Machines \citep{mitsumori2005gene}, Hidden Markov Models \citep{zhou2005recognition}, Maximum Entropy \citep{dingare2005system}, and Conditional Random Fields \citep{settles2004biomedical}. Despite their strengths, these systems required extensive manual adaptation when data changed. Deep learning approaches \citep{habibi2017deep} address this issue by automating feature extraction, though they still rely on large annotated datasets for training      

The introduction of the transformer neural architecture \citep{vaswani_attention_2017} revolutionized natural language processing and artificial intelligence in general. Key components of transformer architecture are its encoder and decoder. The encoder can be trained separately using unsupervised and supervised transfer learning \citep{devlin_bert_2019}, allowing models pre-trained on general text to be fine-tuned for specific tasks with less annotated data. However, base transformer models often underperform in specialized domains \citep{Lee2020}, prompting researchers to pre-train models on biomedical data before fine-tuning for biomedical NER. Notable examples include BioBERT \citep{Lee2020} and SciFive \citep{Phan2021}, which were initialized with BERT and T5 weights, respectively, and further trained on PubMed and PMC texts. Additionally, \citep{gu2021domain} showed that pre-training directly on biomedical data improves downstream performance, leading to PubMedBERT, a BERT model built from scratch on biomedical texts. BioGPT \citep{luo2022biogpt}, a generative transformer pre-trained on biomedical literature, offers an alternative to BioBERT and PubMedBERT but has not been evaluated for NER. In our work, we further explore using an encoder-only transformer for zero-shot NER.

The need to recognize entities in a growing number of scarcely annotated or unannotated texts has led researchers in the direction of zero- and few-shot learning. An approach proposed by \citep{Halder2020} transforms tasks of sentence classification (question type classification, sentiment, and topic classification) into a generic binary classification problem. Feeding the information to the model in the form of a "sentence, label" tuple for each of the tasks and continually fine-tuning the same model for each of the tasks allows the model to make zero-shot predictions, as long as there is enough similarity between tasks. However, TARS faces scalability issues when dealing with a large number of classes - an issue \citep{ziletti2022medical} tackle by proposing xTARS. xTARS combines BERT-based multi-class classification ensembles to deal with scalability problems and improve prediction stability and the TARS zero/few-shot learning approach while making specific choices about the examples shown to the model.

Few-shot learning for general-domain NER is fairly explored \citep{huang2020few}, and while there is some work done on zero-shot learning for general-domain NER \citep{aly2021leveraging, nguyen2021dozen, van2021zero}, there is little work done on few- and zero-shot learning specifically for biomedical NER.  Recently, GPT models and ChatGPT gained popularity for many NLP tasks. Given that GPT models are few-shot learners and can be used in a zero-shot regime, ChatGPT with a specific prompt design was used for a clinical zero-shot approach for clinical NER \citep{hu2023zero}. QaNER system formulated a NER problem as a QA task and tried to approach it with prompt engineering \citep{liu2022qaner}.

\section{Materials and Methods}





In this paper, we introduce a method for biomedical Named Entity Recognition (NER) in English. Our approach involves factorizing the multi-class token classification task into binary token classification. We hypothesize that by turning multi-label classification into binary classification and including the class name in the prompt, we can enable zero- and few-shot named entity classification using encoder models, and better transfer important features between seen labels for classifying new data. This process includes transforming and merging numerous datasets and biomedical entities to fine-tune domain-specific BERT-based language models for performing NER tasks in low-shot scenarios.

Our approach draws inspiration from the text classification method proposed by \citep{Halder2020}, which combines an entity label with the sentence to create the input. For the underlying models, we fine-tuned encoder-based, domain-specific pre-trained language models, such as BioBERT \citep{Lee2020} and PubMedBERT \citep{gu2021domain}. Given our focus on the biomedical domain, utilizing a domain-specific language model is anticipated to yield a performance advantage over general models like BERT.

We propose that the semantic information inherent in named entity labels, along with their potential similarity to labels encountered during pre-training, will enable the model to accurately classify tokens associated with novel labels without requiring explicit examples during fine-tuning. By training the model on a diverse set of named entity classes, our data-driven approach uses transformer architecture to improve the model's ability to understand semantic relationships among these classes. This method, combined with the potential similarity of new labels to those seen during pre-training, enables the model to effectively classify tokens for new labels even without explicit examples during the fine-tuning phase.

\subsection{Dataset}

For fine-tuning a transformer model for the NER task, we needed data already annotated for biomedical named entities. We used six publicly available biomedical corpora which had already been predominantly manually annotated for different named entity classes.

\begin{itemize}

    \item \noindent\textbf{CDR} \citep{wei2016assessing} is a biomedical dataset annotated for chemicals, diseases, and chemical-disease interactions on 1500 PubMed articles. We used exclusively named entity annotations. 

    \item \noindent\textbf{CHEMDNER} \citep{krallinger2015chemdner} consists of 10000 PubMed abstracts annotated with chemical entity mentions. 

    \item \noindent\textbf{BioRED} \citep{luo2022biored} is a biomedical relation extraction dataset with multiple entity types and relation pairs annotated at the document level, on a set of 600 PubMed abstracts. We have used exclusively named entity annotations from this dataset. 

    \item \noindent\textbf{NCBI Disease} \citep{DOGAN20141} is a biomedical dataset annotated with disease mentions, using concept identifiers from either MeSH or OMIM and containing 793 PubMed abstracts.

    \item \noindent\textbf{JNLPBA} \citep{collier2004introduction} is a biomedical dataset that comes from the GENIA version 3.02 corpus created with a controlled search on MEDLINE, consisting of 2404 PubMed abstracts with term annotation.

    \item \noindent\textbf{N2C2} \citep{henry20202018} consists of drug administration-relevant annotations extracted from a total of 1243 de-identified discharge summaries.
\end{itemize}

Except for \textbf{N2C2} corpus, where (medication) information was extracted from de-identified discharge summaries, all 5 other corpora consist of PubMed abstracts and paper titles. While entities across datasets overlap to some degree, they still cover a diverse set of biomedical named entities. 


Given our emphasis on recognizing biomedical named entities, we aimed to encompass a wide spectrum of biomedical subdomains, including drugs, diseases, and chemical compounds, among others. Additionally, we incorporated sub-classes of previously collected classes (e.g., \textit{Specific Disease} for \textit{Disease}, \textit{Protein Type} for \textit{Protein}), enabling exploration of the model's transfer learning behavior from generic to specific classes and vice versa.

In each corpus, all named entities were annotated at the document level, specifically for the classes relevant to the initial purpose of that corpus. Before integrating the corpora, we performed several preprocessing steps to unify them into a single format.

Firstly, documents were split into sentences which were originally labeled for all the named entity classes contained in the specific corpus. Then we performed data transformation, which is described in detail in Section \ref{DataTransform}. After preprocessing, we merged all six datasets into a single dataset, matching identical classes across previously different datasets (e.g., the \textit{Cell Line} class was present in both the BioRED and JNLPBA corpora).

The dataset was subsequently partitioned into training (85\%), validation (5\%), and test (10\%) subsets. These subsets were balanced based on named entity class representation and the count of empty label vectors (without any named entities) versus non-empty label vectors (containing at least one named entity).

\subsubsection{Data transformation method}\label{DataTransform}

In order to employ a binary classification approach, we initially reformatted individual datasets such that each dataset ultimately contained k times as many sentences as at the beginning of the transformation, where k represents the number of named entity classes. Each model input was then formed by concatenating two sequences: the class name and the sentence. The rationale behind this approach was to establish a methodology and dataset wherein each sentence is associated with a specific entity type for annotation. Consequently, the data was structured as follows:


$$<LabelName><Separator><Sentence>$$

where the \textit{LabelName} is a class that is being annotated, and the \textit{Sentence} is a sentence example. The first segment's label name can be considered a prompt defining the named entity class.

The corresponding numerical output vector exclusively includes named entities pertaining to the class indicated in the first segment of the input sequence. Tokens matching the label name from the initial segment are mapped to "1", while all other tokens are mapped to "0".


\begin{figure}[h]
     \centering
     \includegraphics[width=1\linewidth]{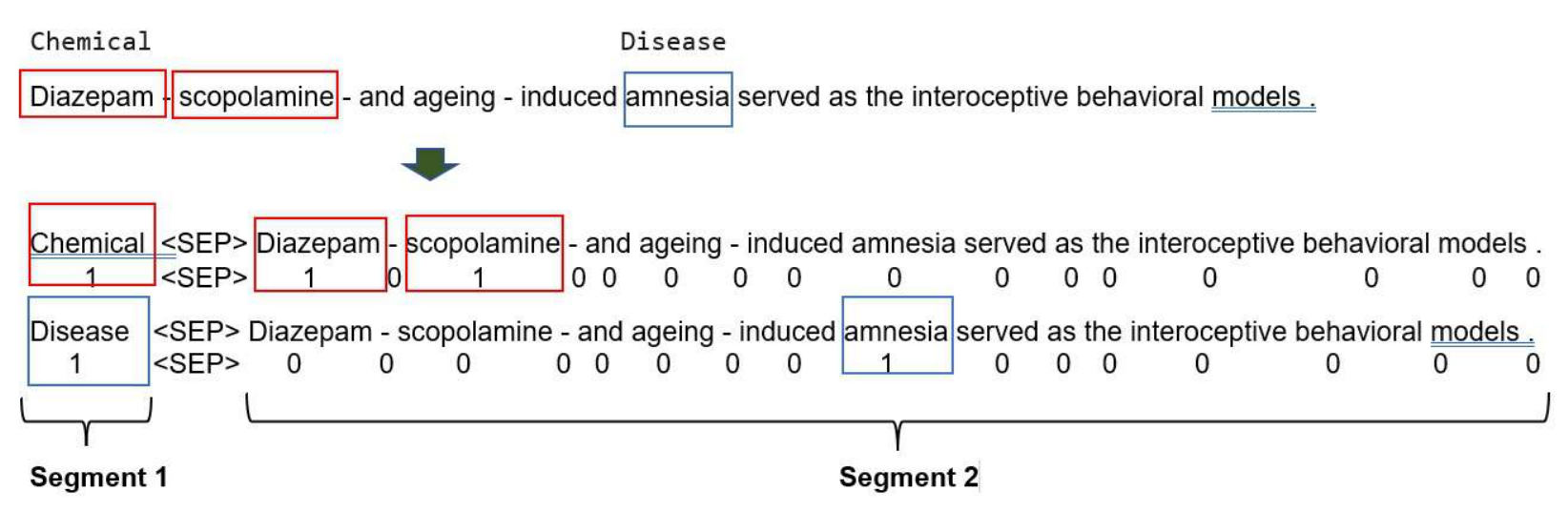}
     \caption{Example of transforming one annotated sentence to the new format of input and expected output.}\label{Fig:InputTransformation}
\end{figure}

This approach enabled us to convert token classification from a multi-class problem into a binary classification task, as depicted in Figure \ref{Fig:InputTransformation}. The example sentence originates from the CDR dataset and encompasses two classes (\textit{Chemical} and \textit{Disease}), consequently, it is transformed into two training sequences.

\subsubsection{Dataset analysis}\label{da}

The final dataset comprises of 708,790 samples (sentences) and encompasses 26 biomedical entity classes. Only 2.67\% of the tokens are labeled as named entities, with the largest class being \textit{Chemical(0.863\%)}, followed by \textit{Protein(0.396\%)} as presented in fourth column of Table \ref{class_token_distribution}. The average sentence length in the newly created merged dataset is 23.2 tokens, with an average of 0.74 labeled tokens per sentence. 

Table \ref{class_token_distribution} presents the number of annotated Named Entities (NE) alongside non-NE tokens and illustrates the distribution of NE across two metrics in the merged dataset: percentage of given class labels in all tokens in dataset and, percentage of given class in all annotated NE in dataset. Table \ref{datasets} details the distribution of classes across different datasets

Among the 26 classes within our merged dataset, some classes denote similar entities but have distinct names, while others denote both similar entities and share similar names. This observation led us to devise two main principles for categorizing all classes into four groups, as delineated in Table \ref{syn-sem-similar}. The syntactic principle determines whether the names of two classes are similar, while the semantic principle discerns whether two classes refer to similar entities.

\begin{table}[!h]
\scriptsize
\centering

\caption{Named Entity class tokens and other tokens distribution metrics for both individual classes and the entire dataset. \label{class_token_distribution}}

\resizebox{\linewidth}{!}{%
\begin{tabular}{|l|l|l|l|l|l|} 
\hline

\makecell[l]{\textbf{Class Name}} & \makecell[l]{\textbf{No. Named} \\ \textbf{Entities}} & \makecell[l]{\textbf{No. of annotated tokens}} &  \makecell[l]{\textbf{Perc. of given class labels} \\ \textbf{in all Dataset tokens (\%)}} & \makecell[l]{\textbf{Perc. of given class label} \\ \textbf{in all annotated NE (\%)}} \\ 
\hline
Chemical & 141555 & 2652627 & 0.863 & 32.28 \\
\hline
Protein & 64958 & 593590 & 0.396 & 14.81 \\
\hline
DNA & 28152 & 593590  & 0.172 & 6.42 \\
\hline
Chemical Family & 21614 & 2302853  & 0.132 & 4.93 \\
\hline
Disease & 21057 & 349774  & 0.128 & 4.80 \\
\hline
Cell Type & 20378 & 593590  & 0.124 & 4.65 \\
\hline
Drug & 16864 & 782232  & 0.103 & 3.85 \\
\hline
Frequency & 14026 & 782348  & 0.086 & 3.20 \\
\hline
Cell Line & 13018 & 609523  & 0.079 & 2.97 \\
\hline
Dosage & 11749 & 782230 & 0.072 & 2.68 \\
\hline
Strength & 11696 & 782230  & 0.071 & 2.67 \\
\hline
Gene Or Gene Product & 10813 & 141289  & 0.066 & 2.47 \\
\hline
Form & 9918 & 782230  & 0.060 & 2.26 \\
\hline
Disease Or Phenotypic Feature & 9132 & 166455 & 0.056 & 2.08 \\
\hline
Specific Disease & 8846 & 169805  & 0.054 & 2.02 \\
\hline
Reason & 6165 & 782240  & 0.038 & 1.41 \\
\hline
Route & 5725 & 782230  & 0.035 & 1.31 \\
\hline
Chemical Entity & 5611 & 116609  & 0.034 & 1.28 \\
\hline
Sequence Variant & 3247 & 68483  & 0.020 & 0.74 \\
\hline
Modifier & 2923 & 130063  & 0.018 & 0.67 \\
\hline
RNA & 2786 & 593590  & 0.017 & 0.64 \\
\hline
Disease Class & 2304 & 101274  & 0.014 & 0.53 \\
\hline
Organism Taxon & 2163 & 153501  & 0.013 & 0.49 \\
\hline
Duration & 1480 & 782230  & 0.009 & 0.34 \\
\hline
ADE & 1459 & 782232  & 0.009 & 0.33 \\
\hline
Composite Mention & 922 & 27319  & 0.006 & 0.21 \\
\hline
\end{tabular}}
\end{table}

\begin{table}[!h]
\scriptsize
\centering
\caption{Class and token distribution in individual data sets. Column A: Number of sentences containing the class, Column B: Number of all tokens in sentences with a prefix of a given class, Column C: Number of annotated named entity tokens per class. The samples were created on a document level (examples were created with all classes that were annotated in the given document for all sentences), with short sentences being filtered out (less than 3 words) \label{datasets}}

\resizebox{\linewidth}{!}{%
\begin{tabular}{|l|l|l|l|l|} 
\hline
~                          & \textbf{Class}~                         & \textbf{A}~ & \textbf{B}~ & \textbf{C}~  \\ 
\hline
\multirow{4}{*}{NCBI~}     & \textbf{Specific Disease}~              & 7029~                              & 169805~                   & 8846 (5.39\%)~~                                                       \\ 
\cline{2-5}
                           & \textbf{Composite Mention}~             & 1119~                              & 27319~                    & 922 (3.37\%)~~                                                       \\ 
\cline{2-5}
                           & \textbf{Modifier}~                      & 5343~                              & 130063~                   & 2923 (2.26\%)~~                                                       \\ 
\cline{2-5}
                           & \textbf{Disease Class}~                 & 4158~                              & 101274~                   & 2304 (2.23\%)~~                                                       \\ 
\hline
\multirow{6}{*}{BIORED~}   & \textbf{Sequence Variant}~              & 3246~                              & 68483~                   & 3247 (6.16\%)~~                                                       \\ 
\cline{2-5}
                           & \textbf{Gene Or Gene Product}~          & 6194~                              & 141289~                  & 10813 (6.08\%)~~                                                       \\ 
\cline{2-5}
                           & \textbf{Disease Or Phenotypic Feature}~ & 7599~                              & 166455~                   & 9132 (5.15\%)~~                                                       \\ 
\cline{2-5}
                           & \textbf{Chemical Entity}~               & 5337~                              & 116609~                   & 5611 (4.42\%)~~                                                       \\ 
\cline{2-5}
                           & \textbf{Cell Line}~                     & 601~                               & 15933~                    & 312 (1.58\%)~~                                                       \\ 
\cline{2-5}
                           & \textbf{Organism Taxon}~                & 6923~                              & 153501~                   & 2163 (1.35\%)~~                                                       \\ 
\hline
\multirow{2}{*}{CDR~}      & \textbf{Disease}~                       & 17015~                             & 349774~                  & 21057 (5.89\%)~~                                                       \\ 
\cline{2-5}
                           & \textbf{Chemical}~                      & 17015~                             & 349774~                  & 20366 (5.38\%)~~                                                       \\ 
\hline
\multirow{2}{*}{CHEMDNER~} & \textbf{Chemical}~                      & 95221~                             & 2302853~                 & 121189 (4.11\%)~~                                                       \\ 
\cline{2-5}
                           & \textbf{Chemical Family}~               & 95221~                             & 2302853~                  & 21614 (0.81\%)~~                                                       \\ 
\hline
\multirow{5}{*}{JNLPBA~}   & \textbf{Protein}~                       & 22402~                             & 593590~                  & 64958 (10.98\%)~~                                                      \\ 
\cline{2-5}
                           & \textbf{DNA}~                           & 22402~                             & 593590~                  & 28152 (4.53\%)~~                                                       \\ 
\cline{2-5}
                           & \textbf{Cell Type}~                     & 22402~                             & 593590~                  & 20378 (3.57\%)~~                                                       \\ 
\cline{2-5}
                           & \textbf{Cell Line}~                     & 22402~                             & 593590~                  & 12706 (2.05\%)~~                                                       \\ 
\cline{2-5}
                           & \textbf{RNA}~                           & 22402~                             & 593590~                   & 2786 (0.46\%)~~                                                       \\ 
\hline
\multirow{9}{*}{n2c2~}     & \textbf{Drug}~                          & 36078~                             & 782232~                  & 16864 (2.40\%)~~                                                       \\ 
\cline{2-5}
                           & \textbf{Frequency}~                     & 36136~                             & 782348~                  & 14026 (2.01\%)~~                                                       \\ 
\cline{2-5}
                           & \textbf{Strength}~                      & 36077~                             & 782230~                  & 11696 (1.52\%)~~                                                       \\ 
\cline{2-5}
                           & \textbf{Dosage}~                        & 36077~                             & 782230~                  & 11749 (1.50\%)~~                                                       \\ 
\cline{2-5}
                           & \textbf{Form}~                          & 36077~                             & 782230~                   & 9918 (1.40\%)~~                                                       \\ 
\cline{2-5}
                           & \textbf{Reason}~                        & 36082~                             & 782240~                   & 6165 (0.88\%)~~                                                       \\ 
\cline{2-5}
                           & \textbf{Route}~                         & 36077~                             & 782230~                   & 5725 (0.85\%)~~                                                       \\ 
\cline{2-5}
                           & \textbf{ADE}~                           & 36078~                             & 782232~                   & 1459 (0.19\%)~~                                                       \\ 
\cline{2-5}
                           & \textbf{Duration}~                      & 36077~                             & 782230~                   & 1480 (0.16\%)~                                                        \\
\hline
\end{tabular}
}
\end{table}

\begin{table*}[!h]
\centering
\caption{Examples of semantically and syntactically (dis)similar classes in the merged dataset.}
\label{syn-sem-similar}
\begin{tabular}{|ll|cc|}
\hline
\multicolumn{2}{|c|}{\multirow{2}{*}{}} & \multicolumn{2}{c|}{Syntactic}                                        \\
\multicolumn{2}{|c|}{}                  & similar                                          & dissimilar         \\ \hline
\multirow{2}{*}{Semantic}  & similar    & \multicolumn{1}{l|}{\textit{Disease} : \textit{Specific Disease}} & \textit{Chemical} : \textit{Drug}   \\ \cline{3-4} 
                           & dissimilar & \multicolumn{1}{l|}{\textit{Cell Line} : \textit{Cell Type}}      & other combinations \\ \hline
\end{tabular}
\end{table*}

\subsection{Model architecture and training method}

\begin{figure}[h]
\centering
\includegraphics[width=0.9\linewidth]{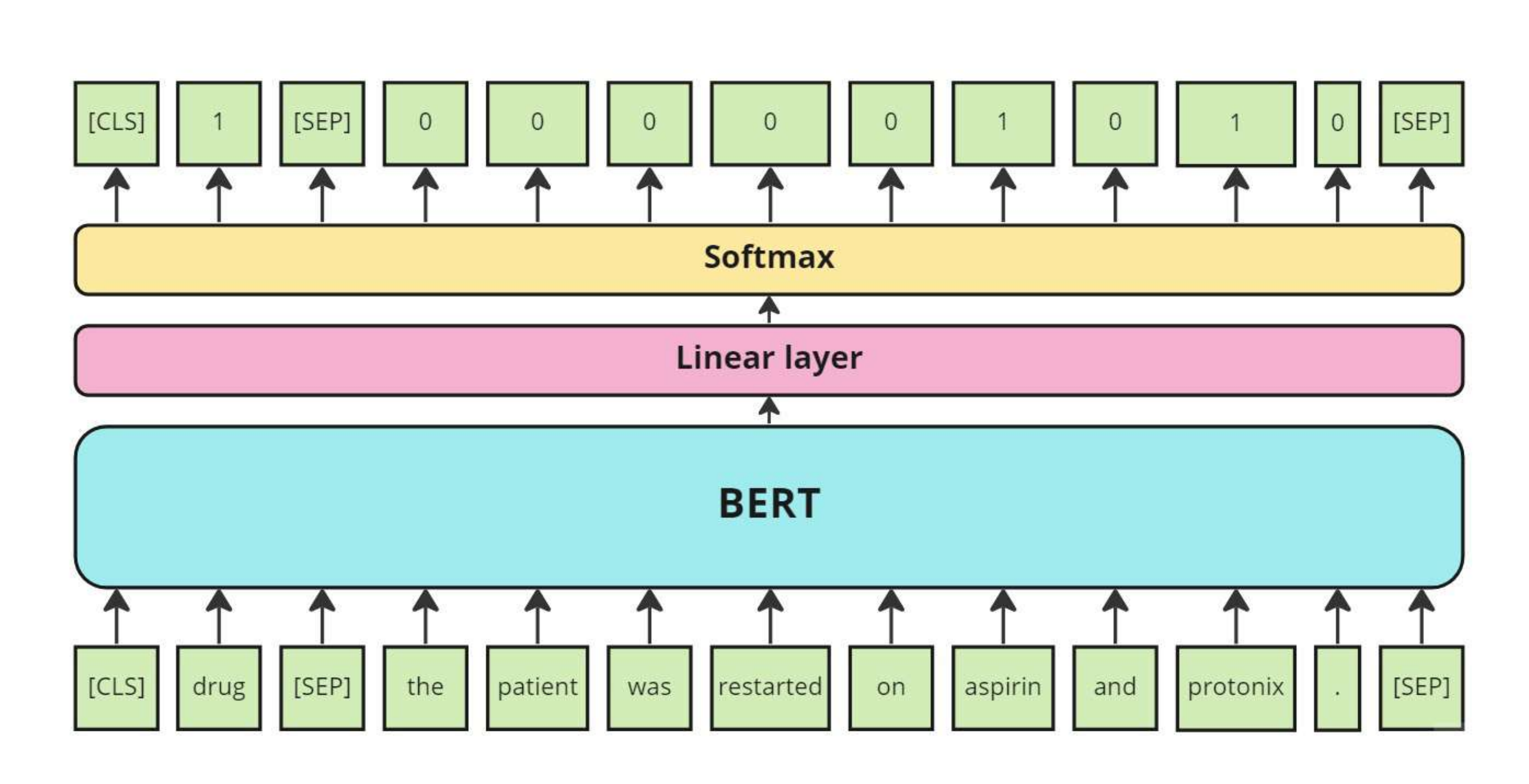}
\caption{Model architecture (BERT refers to fine-tuned BioBERT or PubmedBERT models)}
\label{BERT}
\end{figure}

Our model utilizes a BERT architecture, modified for Named Entity Recognition (NER) by incorporating one linear and one softmax layer for token classification at the top of the base model. The model discerns each token's classification as a named entity or non-entity (refer to Figure \ref{BERT}).

The BERT architecture is pre-trained on both the next sentence prediction task and masked language modeling \citep{devlin_bert_2019}. For the next sentence prediction task, the input text is divided into two segments, allowing self-attention layers to learn dependencies and relationships among tokens across segments. Leveraging this capability, we position the label for the named entity in the first segment and the sentence to be labeled in the second segment of each input (see Figure \ref{Fig:InputTransformation}). We hypothesize that our model will learn the desired graph connections between different segments during the fine-tuning process. Supervision is established by assigning the same class label (in our case, "1") to tokens in the second segment corresponding to the occurrence of the named entity defined in the first segment for seen classes, while all other tokens are labeled as zeros. This process is illustrated in Figure~\ref{BERT}.

We employed two transformer models for fine-tuning on the NER task with our dataset: BioBERT (biobert-v1.1), a model pre-trained on texts from the general domain and fine-tuned for the biomedical domain; and PubMedBERT (PubMedBERT-base-uncased), a model pre-trained on biomedical data from scratch. The inputs for these models were initially tokenized using the respective tokenizer. It is important to note that tokenizers for BioBERT and PubMedBERT differ significantly, as explained in detail in \cite{gu2021domain}.

Our primary objective was to train the model on all 26 classes from our transformed dataset to enable it for the NER task on novel named entity classes. To avoid annotating new classes for testing the model's performance, we opted to train our model on 25 named entity classes and reserve the 26th class, unseen by the model, for evaluating its performance. As an illustrative example of the model's behavior when encountering unseen named entity classes, we selected 9 classes, and for each of them, we fine-tuned base transformer models in two stages:

\begin{itemize}
    \item \textbf{Initial fine-tuning or pretraining} - For each of the 9 chosen classes, we trained a separate model on the entire dataset, excluding only one (unseen) of the selected classes at a time to evaluate the performance of the Initial model in zero-shot mode. These models should be capable of recognizing named entities from all 25 pre-trained classes, as well as semantically similar unseen class. To create the training subset for each named entity class, examples of that class were removed from the initial training subset. The validation set remained unchanged, containing all classes. The examples of the unseen class were selected from the initial test subset for testing the model's performance. Initial training utilized a batch size of 32 and trained the model for 6 epochs.
    \item \textbf{Few-shot fine-tuning} - Following Initial fine-tuning for a specific class, we further fine-tuned that model with 1, 10, and 100 supporting examples of the unseen class. These examples were drawn from the pool of examples previously removed from the training subset during the Initial fine-tuning process. Model performance was evaluated using selected unseen class examples from the test subset. For 1-, 10-, and 100-shot training, a batch size of 1 was employed, and the model was trained for 10 epochs.
\end{itemize}

We conducted this procedure for all 9 selected classes, considering the semantic and syntactic relations between them. Namely, we have selected classes that are both semantically and syntactically similar (Disease Specific Disease), semantically similar classes, but syntactically different (Drug - Chemical, RNA - DNA), and classes that are both semantically and syntactically distant from one another (e.g. Cell Line, Dosage, Protein).

All trainings utilized the ADAM optimizer \citep{kingma2014adam} with a learning rate of 5e-5 and weight decay of 0.01. Training sessions were executed on a single DGX NVIDIA A100-40GB GPU using PyTorch and the Hugging Face transformer library. Model testing was performed by employing joined token-level partial similarity, assessing standard measures of F1 score, accuracy, precision, and recall.




\section{Experiments and results}

To facilitate the discovery of new named entity classes with minimal or no additional examples, we devised a solution by decomposing multi-class token classification into binary classification. This approach allows us to include the class name as a descriptor in the input, enabling the identification of previously encountered and unseen entities in the text. To ascertain the viability of factorizing multi-class token classification into binary token classification, we conducted an initial pilot study, followed by the main experiments.

\subsection{Pilot study: Comparison of binary and multi-class NER}

\begin{figure}[!h]
     \centering
     \includegraphics[width=0.9\linewidth]{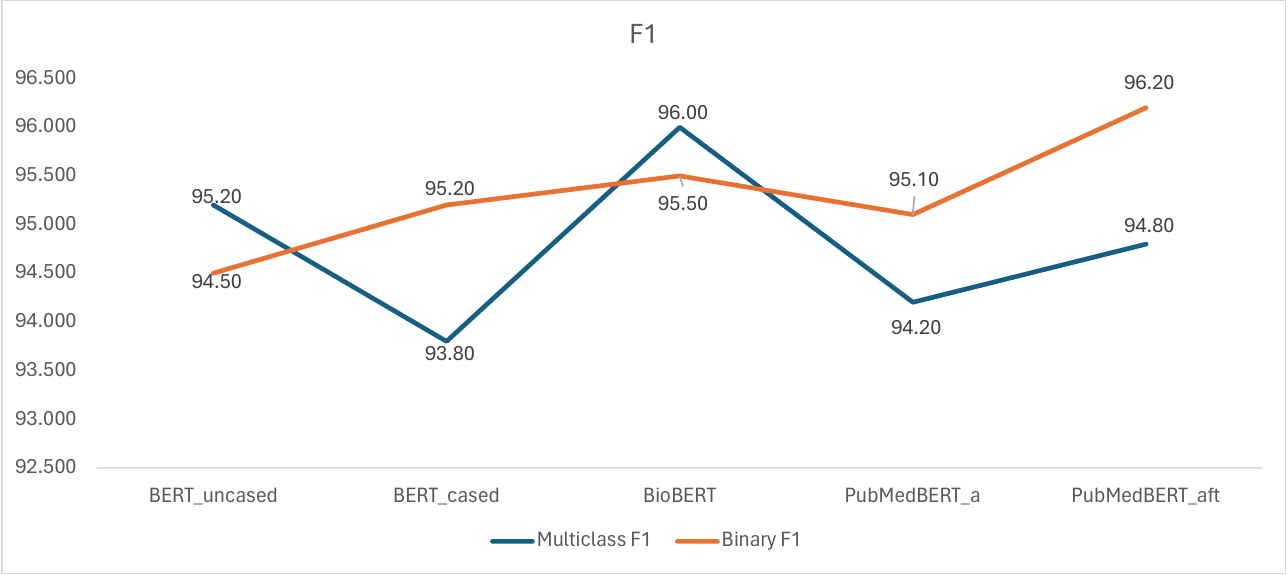}
     \caption{Comparison of average F1 scores for the NER task on BC5CDR corpus using five fine-tuned BERT-based models for the tasks of binary classification and multi-class NER.}\label{Fig:pilotGraph}
\end{figure}

We conducted a comparative analysis of multi-class and binary classification on the BC5CDR dataset \cite{wei2016assessing} using five transformer models: (1) BERT\_uncased and (2) BERT\_cased, pre-trained on English Wikipedia and BooksCorpus; (3) BioBERT, a general BERT model fine-tuned on PubMed abstracts and full-text articles; and finally, (4) PubMedBERT\_a and (5) PubMedBERT\_aft, pre-trained from scratch on PubMed abstracts and PubMed abstracts and full-text articles, respectively. These models were selected to represent a spectrum ranging from those trained solely on general texts, through a combination with domain texts, to purely domain-specific texts.

The results, depicted in Figure \ref{Fig:pilotGraph}, demonstrate that the task of multi-class token classification can be effectively factorized into binary classification. Notably, the results for the NER categories show similarities, with binary token classification often exhibiting favorable outcomes.

\subsection{Main study: Zero-shot and few-shot NER using binary token classification}

Initially, we conducted a comparison in a low-shot regime between the bare BioBERT model (without fine-tuning on our dataset) and our fine-tuned version of BioBERT using the dataset we generated. Our findings indicate that our approach yields superior results. Figures \ref{baseline} and \ref{fine-tuned} depict the performance comparison for BioBERT across 2 classes.

Next, we compared the performance of two models fine-tuned with our transformed dataset. Fine-tuning the base models of BioBERT and PubMedBERT yielded similar results. However, PubMedBERT, being more domain-specific, showed an average improvement of 8.3\% for zero-shot (Figure \ref{0-shot}) and 3.4\% for 100-shot learning (Figure \ref{100-shot}) compared to BioBERT.

Table \ref{Result-FirstSegment1} shows the test results for 9 classes using BioBERT and PubMedBERT models, along with the best epoch for each class, determined by our validation method. The analysis of these results is based on the semantic and syntactic similarity between classes (see Section \ref{da}).

In the 0- to 10-shot regimes, the highest F1 score was achieved for the class \textit{Disease}, which also ranked among the highest in the 100-shot regime. This outcome was anticipated, as \textit{Disease} exhibits the most significant number of direct or indirect semantic connections to other classes in our dataset (refer to Table \ref{datasets}). Notably, \textit{Specific Disease}, being syntactically and semantically akin to \textit{Disease}, demonstrated promising results in the 0-shot regime, indicating a successful transfer learning process between these two classes.

\textit{Chemical} exhibits the highest ratio of annotated entities in the dataset, significantly facilitating knowledge transfer to semantically similar classes (such as \textit{Drug}). However, the transfer from other semantically similar classes to \textit{Chemical} was comparatively lower in the 0-shot regime. This disparity may be attributed to the fact that other classes semantically akin to \textit{Chemical} serve as its semantic subordinates, thereby diminishing their contribution to the transfer learning process.

In contrast, \textit{Dosage}, being semantically and syntactically independent of other classes, registered the lowest score in the 0-shot regime, with no examples retrieved. Nonetheless, its performance improves with a few supporting examples, occasionally surpassing the results for mutually semantically similar classes. This is owing to the specific structure of \textit{Dosage}, typically comprising measures such as mg, g, ml, etc., facilitating seamless learning based on a limited number of examples.

Most of the classes saturate between 77-87\% F1 score after 100 shots, depending on the entity, which is not far from the reported state-of-the-art NER results trained on the whole dataset (for example, \citep{kuhnel2022we} reported an F1 score of 87.27\% for NCBI corpus and 83.07\% for CDR corpus using BioBERT, BioRED reported F1-score of 89.3\% \citep{luo2022biored}).


\begin{table}[b]
\centering
\begin{tabular}{@{}c@{}c@{}}
    \includegraphics[width=0.5\textwidth]{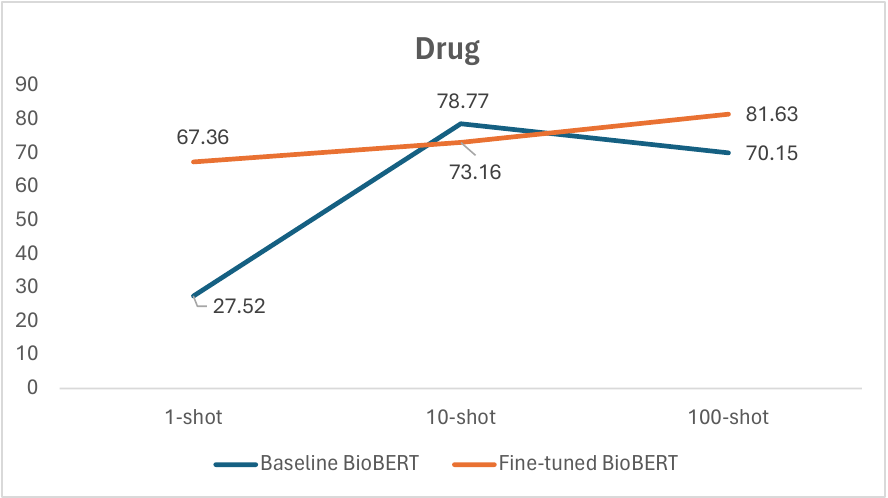} &
    \includegraphics[width=0.5\textwidth]{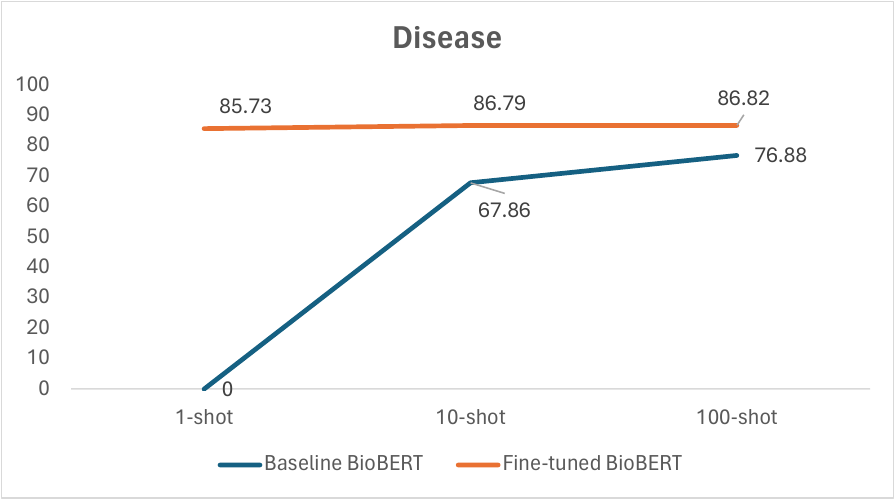} \\
    \parbox{0.5\textwidth}{
      \captionsetup{justification=raggedright, singlelinecheck=false, format=plain}
      \captionof{figure}{Class \textit{Drug} F1 comparison between baseline and\\ fine-tuned BioBERT.}
      \label{baseline}
    } &
    \parbox{0.5\textwidth}{
      \captionsetup{justification=raggedright, singlelinecheck=false, format=plain}
      \captionof{figure}{Class \textit{Disease} F1 comparison between baseline and\\fine-tuned BioBERT.}
      \label{fine-tuned}
    }
\end{tabular}
\end{table}

\begin{table}[!h]
\centering
\begin{tabular}{@{}c@{}c@{}}
    \includegraphics[width=0.5\textwidth]{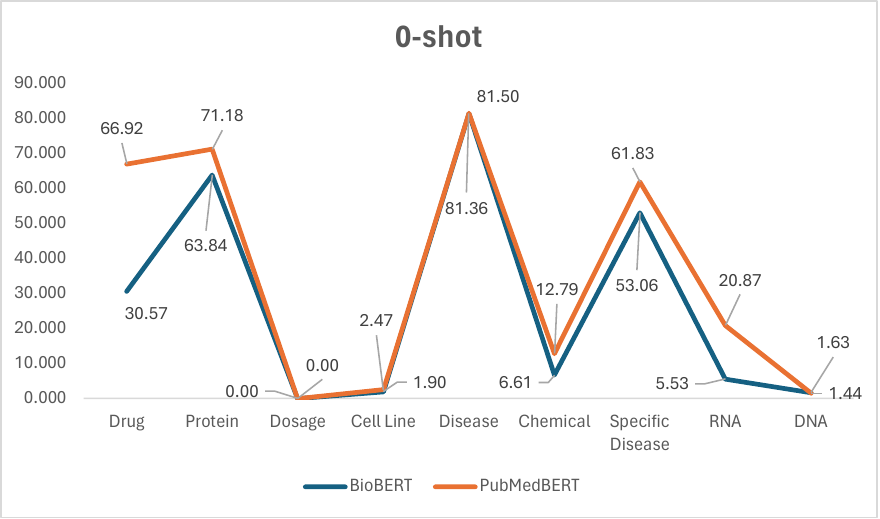} &
    \includegraphics[width=0.5\textwidth]{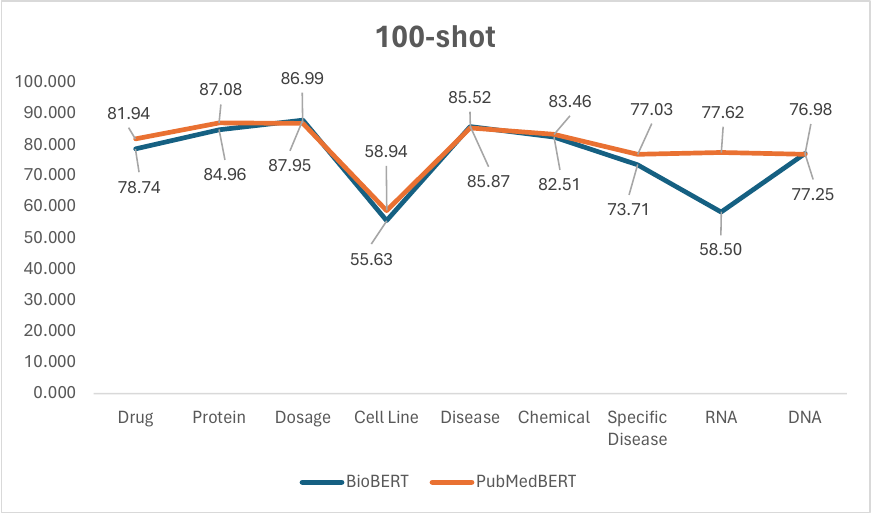} \\
    \parbox{0.5\textwidth}{
      \captionsetup{justification=raggedright, singlelinecheck=false, format=plain}
      \captionof{figure}{0-shot F1 score comparison between BioBERT and \\PubMedBERT.}\label{0-shot}
    } &
    \parbox{0.5\textwidth}{
      \captionsetup{justification=raggedright, singlelinecheck=false, format=plain}
      \captionof{figure}{100-shot F1 score comparison between BioBERT and \\PubMedBERT.}\label{100-shot}
    }
\end{tabular}
\end{table}

\newcommand{\STAB}[1]{\begin{tabular}{@{}c@{}}#1\end{tabular}}

\begin{table}[t]
\scriptsize
\centering
\caption{F1 (precision, recall) scores given in percentages and best epoch for BioBERT and PubMedBERT. The column with zero shows the test results of a specific unseen class on the Initial model by training without any examples of that class (0-shot). The following columns present the results of testing the fine-tuned Initial model with 1, 10, and 100 examples of a previously unseen class (1-shot, 10-shot and 100-shot). \label{Result-FirstSegment1}}
\centering

\begin{tabular}{|c|l|c|l|l|l|l|}
\cline{1-7}
\multicolumn{1}{|l|}{\multirow{2}{*}{\textbf{}}} & \multicolumn{1}{c|}{\multirow{2}{*}{\textbf{Class}}} & \multicolumn{1}{c|}{\multirow{2}{*}{\textbf{Epoch}}} & \multicolumn{4}{|c|}{\textbf{Number of supporting examples }}                                                                                                                                                                                                                                                                                                                                    \\ \cline{4-7} 
\multicolumn{1}{|l|}{}                                & \multicolumn{1}{c|}{}
& \multicolumn{1}{c|}{}
& \multicolumn{1}{c|}{0}                                                                          & \multicolumn{1}{c|}{1}                                                                         & \multicolumn{1}{c|}{10}                                                                        & \multicolumn{1}{c|}{100}                                                                       \\ \cline{1-7}
\multirow{9}{*}{\STAB{\rotatebox[origin=c]{90}{\textbf{BioBERT}}}}
& Drug & 2                                                & 30.57 \tiny(79.41,18.93)          & 79.83 \tiny(82.37,77.44)          & 78.00 \tiny(76.41,79.67)          & 78.74 \tiny(71.69,87.32)          \\
                                                     & Protein & 2                                             & 63.84 \tiny(76.13,54.96)          & 71.74 \tiny(71.13,72.36)          & 77.51 \tiny(78.13,76.89)          & 84.96 \tiny(79.41,91.35)          \\
                                                     & Dosage  & 2                                             & 0.00 \tiny(0.00,0.00)             & 0.15 \tiny(0.91,0.08)             & 78.03 \tiny(76.73,79.38)          & \textbf{87.95} \tiny(83.88,92.42)          \\
                                                     & Cell Line   & 3                                         & 1.90 \tiny(1.61,2.33)             & 3.49 \tiny(2.85,4.50)             & 48.54 \tiny(33.25,89.87)          & 55.63 \tiny(40.49,88.82)          \\
                                                     & Disease     & 3                                         & \textbf{81.36} \tiny(87.61,75.94) & \textbf{86.22} \tiny(84.99,87.48) & \textbf{86.72} \tiny(82.68,91.19) & 85.87 \tiny(82.90,89.06) \\
                                                     & Chemical    & 1                                         & 6.61 \tiny(19.26,3.99)            & 73.89 \tiny(66.86,82.58)          & 73.63 \tiny(74.47,72.79)          & 82.51 \tiny(80.40,84.74)          \\
                                                     & Specific Disease      & 1                               & 53.06 \tiny(48.58,58.45)          & 60.08 \tiny(51.76,71.60)          & 63.32 \tiny(50.94,83.65)          & 73.71 \tiny(64.77,85.52)          \\
                                                     & RNA    & 1                                              & 5.53 \tiny(3.09,26.03)            & 14.38 \tiny(7.78,94.52)           & 13.44 \tiny(7.24,92.81)           & 58.50 \tiny(45.30,82.53)          \\
                                                     & DNA    & 2                                              & 1.63 \tiny(7.81,0.91)             & 18.64 \tiny(16.01,22.29)          & 78.56 \tiny(71.25,87.55)          & 77.25 \tiny(67.18,90.89)          \\ \cline{2-7}
                                                     & Average   & 2                                                 & 27.17 \tiny(35.94,26.84)	& 45.38 \tiny(42.74,	56.98) &	66.42 \tiny(61.23,83.75)		& 76.12 \tiny(68.45,88.07)           \\ \hline
\multirow{9}{*}{\STAB{\rotatebox[origin=c]{90}{\textbf{PubMedBERT}}}}
& Drug    & 2                                                & 66.92 \tiny(73.20,61.63)          & 77.73 \tiny(72.67,83.56)          & 80.10 \tiny(77.74,82.60)          & 81.94 \tiny(75.57,89.48)          \\
                                                     & Protein    & 3                                             & 71.18 \tiny(83.13,62.23)          & 73.85 \tiny(73.65,74.05)          & 80.80 \tiny(82.10,79.53)          & \textbf{87.08} \tiny(84.29,90.06)          \\
                                                     & Dosage       & 3                                           & 0.00 \tiny(0.00,0.00)             & 52.42 \tiny(45.11,62.54)          & 85.67 \tiny(95.36,77.78)          & 86.99 \tiny(82.91,91.50) \\
                                                     & Cell Line     & 3                                          & 2.47 \tiny(2.10,3.00)             & 3.54 \tiny(2.92,4.50)             & 64.10 \tiny(51.91,83.80)          & 58.94 \tiny(43.55,91.15)          \\
                                                     & Disease     & 2                                            & \textbf{81.50} \tiny(91.65,73.37) & \textbf{85.19} \tiny(85.09,85.30) & \textbf{87.28} \tiny(84.38,90.40) & 85.52 \tiny(80.39,91.34)          \\
                                                     & Chemical     & 1                                           & 12.79 \tiny(32.08,7.99)           & 74.96 \tiny(68.66,82.54)          & 75.37 \tiny(75.45,75.29)          & 83.46 \tiny(79.35,88.02)          \\
                                                     & Specific Disease     & 2                                   & 61.83 \tiny(53.66,72.93)          & 60.94 \tiny(52.16,73.26)          & 65.70 \tiny(53.43,85.30)          & 77.03 \tiny(68.59,87.85)          \\
                                                     & RNA     & 2                                                & 20.87 \tiny(12.18,72.60)          & 17.39 \tiny(9.72,82.53)           & 14.59 \tiny(7.90,95.21)           & 77.62 \tiny(73.25,82.53)          \\
                                                     & DNA    & 2            
                                                     
                                                     & 1.44 \tiny(12.65,0.76)            & 4.86 \tiny(4.44,5.37)             & 75.87 \tiny(66.73,87.91)          & 76.98 \tiny(65.42,93.50)  \\
                                                 \cline{2-7}
                                                     & Average    & 2                                            & \textbf{35.44} \tiny(40.07,39.39)           & \textbf{50.10} \tiny(46.05,61.52)            & \textbf{69.94} \tiny(66.11,	84.20)        & \textbf{79.51}  \tiny(72.59,89.49)\\
                                                     		
                                                \cline{1-7}

\end{tabular}
\end{table}

\section{Discussion}
\label{Discussion}

We have introduced a promising approach for zero- and few-shot NER for biomedical entities The main novelty of our approach i is transforming the input to generate BERT-compatible prompts for zero- and few-shot bimedical NER. . We demonstrate that the transformer network is able to learn and apply the given input to unseen classes.  Additionally, pretraining a model on multiple datasets with a larger number of named entity classes contributes to successful zero- and few-shot NER.

The macro-average F1 score for zero-shot NER using the PubMedBERT model across the 9 tested classes is 35.44\%. If we exclude the class \textit{Dosage}, which exhibited markedly different performance compared to other classes, the macro-average score increases to nearly 40\%. Notably, our method outperforms many reported results in general domain zero-shot learning literature. For instance, \citep{van2021zero} and \citep{aly2021leveraging} reported macro-average F1-scores of 23\% and \citep{nguyen2021dozen} reported a macro-average F1-score of up to 30\% for zero-shot learning. Our method in the zero-shot regime significantly surpasses most of these approaches.

To the best of our knowledge, even significantly larger and more complex proprietary chatGPT models (e.g. GPT-3.5 based models) have reported similar performance, with an F1 score of 41.8\% for exact Named Entity Recognition (NER) matching in clinical NER \citep{hu2023zero}. The models utilized for ChatGPT are approximately 1000 times larger and considerably slower compared to the models we employed (110 million parameters in BERT models, contrasted with 175 billion parameters in the GPT 3.5 model used in chatGPT). At present, hosting this proprietary model on consumer electronics is not feasible due to its extensive size, resulting in slow performance and significant costs associated with hosting.

In terms of performance, we observed three types of behaviors across entity classes:

\begin{itemize}
    \item \textbf{The first pattern} can be seen in entities such as \textit{Dosage}, \textit{RNA} or \textit{DNA}. These classes exhibit poor performance in zero-shot (between 0\% and 20\%), but are steeply learned based on a few supporting examples, and reach 77-87\% F1 score for 100-shot. Poor zero-shot performance is attributed to the semantic independence of these classes from the seen classes during zero-shot fine-tuning. However, these classes appear in a very specific context that the algorithm can learn fast from during few-shot fine-tuning. For example, class \textit{Dosage} is often followed by quantity information, such as "milligram".
    \item \textbf{The second pattern} is observed in entities such as \textit{Drug}, \textit{Protein} and \textit{Specific Disease} where zero-shot performance is fairly good (over 66\% on average), while the performance between 1- and 100-shot shows a mild incline, reaching over 82\% on average. This pattern is attributed to the semantic similarity between unseen classes and the classes seen during zero-shot fine-tuning. For instance, classes \textit{Drug} and \textit{Protein} are semantically subordinate to the seen class \textit{Chemical}, which could be considered their hypernym.
    \item \textbf{The third pattern} is exemplified by the class \textit{Disease} where excellent zero-shot performance is obtained (81.5\%), followed by a very mild increase towards 100-shot performance (85.5\%). This pattern could be explained by the fact that class \textit{Disease} is both syntactically and semantically related to the seen class \textit{Specific Disease} (as its hypernym).
\end{itemize}

It's crucial to highlight that our model, based on the mentioned dataset and trained on 26 biomedical entities, is capable of recognizing any of these entities with performance close to state-of-the-art results reported in \citep{kuhnel2022we} and \citep{luo2022biored}. Moreover, the model can identify new entities that it has never encountered before. In such cases, the model seeks similarity between any semantically similar new class and some of the previously trained classes. Furthermore, the model can be fine-tuned with a few supporting examples of the new class, as demonstrated, leading to significant improvements in the results for the new class.

\subsection*{Selecting the model during zero-shot fine-tuning} \label{Validation}

During our experiments, we encountered challenges in selecting the appropriate validation method for choosing the best-performing zero-shot base model, as well as observed instability in zero-shot results across different runs. In the literature, there is no consensus on the appropriate method for validating and selecting a zero-shot model. We have considered five options:

\begin{itemize}
    \item \textit{Create a validation dataset containing all of the classes, including the unseen class} - during the validation we consider performance on all the classes. Seen classes would represent the majority of the validation dataset and the performance of unseen class may be underrepresented, since the focus is on keeping the model performing well on all the classes while also performing on unseen class. 
    \item \textit{Create a validation dataset containing all of the seen classes, excluding the one we want to evaluate for zero-shot performance} - validation focuses on the performance of seen classes. As it is not possible to know what will novel classes be in the future, this kind of validation is closest to a real use case. 
    \item \textit{Create a validation dataset containing only a single unseen class} - validation focuses on a single unseen class and adjusts the performance towards it. This may lead to optimization around the performance of this unseen class. 
    \item \textit{Create a validation dataset containing a few unseen classes} - similar to the previous case, while more realistic, and less prone to overoptimization to the selected unseen classes, however, it is deviating from evaluation best practices. Testing of the zero-shot method should be done without providing a tested class in the validation set.  
    \item \textit{Manual inspection of various models occurring throughout zero-shot fine-tuning process} - validation is manual. While we can ensure there are no unseen classes affecting training, manual model inspection and evaluation is time-consuming and expensive. 
\end{itemize}

These approaches range from validating solely on unseen classes, to including both seen and unseen classes in validation, to using only seen classes for validation. Each approach has its theoretical advantages and practical limitations.

During the model validation, we used a subset containing both seen classes and an unseen class. Our motivation in doing so was, that the validation is still primarily done on seen classes and that the model will generalize to any unseen class. However, including an unseen class in the validation subset may help in selecting the model's epoch.

This validation technique may not always be feasible in practical scenarios. In many cases, it is preferable to train and validate the dataset solely on seen classes, as the unseen class during deployment is unknown. Unfortunately, rerunning all the models with a new validation subset containing only seen classes was logistically impractical, given that fine-tuning of the base model took over 7 hours per run. As a compromise, we present a comparison of the results for three different classes based on two validation sets (with and without unseen classes) using the BioBERT model in Table \ref{Result-FirstSegment1-valid}.

\begin{table}[!th]
\scriptsize
\centering
\caption{Comparison of validation on BioBERT with 1 on the first segment with and without seen class in the validation set. Results are given as F1 score (precision, recall). \label{Result-FirstSegment1-valid}}
\centering

\begin{tabular}{|c|l|l|l|l|l|}
\cline{1-6}
\multicolumn{1}{|l|}{\multirow{2}{*}{\textbf{Model}}} & \multicolumn{1}{c|}{\multirow{2}{*}{\textbf{Class}}} & \multicolumn{4}{c|}{\textbf{Number of supporting examples}}                                                                                                                                                                                                                                                                                                                                    \\ \cline{3-6} 
\multicolumn{1}{|l|}{}                                & \multicolumn{1}{c|}{}                                & \multicolumn{1}{c|}{0}                                                                          & \multicolumn{1}{c|}{1}                                                                         & \multicolumn{1}{c|}{10}                                                                        & \multicolumn{1}{c|}{100}                                                                       \\ \hline
\multirow{3}{*}{\STAB{\textbf{Without unseen}}}
& Chemical                                                 & 2.10 \tiny(6.83,1.24)          & 25.15 \tiny(45.30,17.41)          & 72.33 \tiny(69.04,75.95)          & 78.21 \tiny(67.84,92.33)          \\
        & Protein                                              & 49.18 \tiny(78.46,35.81)          & 74.28 \tiny(77.42,71.39)          & 80.02 \tiny(82.77,77.44)          & 86.05 \tiny(80.92,91.89)          \\
         & Disease                                               & 81.36 \tiny(87.61,75.94)             & 86.22 \tiny(84.99,87.48)             & 87.16 \tiny(83.98,90.59)          & 85.24 \tiny(80.02,91.19) 
                                                              \\ \hline
\multirow{3}{*}{\STAB{\textbf{With unseen}}}
& Chemical   & 6.61 \tiny(19.26,3.99)          & 73.89 \tiny(66.86,82.58)          & 73.63 \tiny(74.47,72.79)          & 82.51 \tiny(80.40,84.74)          \\
        & Protein   & 63.84 \tiny(76.13,54.96)          & 71.74 \tiny(71.13,72.36)          & 77.51 \tiny(78.13,76.89)          & 84.96 \tiny(79.41,91.35)          \\
         & Disease                                               & 81.36 \tiny(87.61,75.94)             & 86.22 \tiny(84.99,87.48)             & 86.72 \tiny(82.68,91.19)          & 85.87 \tiny(82.90,89.06) 
                                                              \\ \hline

\end{tabular}
\end{table}

The three classes selected for comparison are representatives of the three performance patterns observed in Section \ref{Discussion}, confirming the tendencies of these patterns. 
We observe that \textit{Chemical}, a representative of the first pattern (poor zero-shot performance and steep improvement with the addition of supporting examples), demonstrates better results when validated on both seen and unseen classes for all shots, particularly for zero-shot. On the other hand, \textit{Protein}, a representative of the second pattern (showing mild improvement with the addition of supporting examples), exhibits better performance when having unseen class examples in the validation set for zero-shot. However, for few-shot learning, it shows improvement when validated on a subset containing only seen class examples. Lastly, \textit{Disease}, a representative of the third pattern (having almost constant results from zero to a hundred supporting examples), behaves the same regardless of the validation technique used.

These results show that the selection of the model can be just slightly influenced or improved by adding an unseen class to the validation subset. However, both validation techniques yield comparable results. Therefore, we conclude that for practical deployment purposes, the model can be validated using only examples from seen classes - the ones utilized for the base model fine-tuning.

\section{Limitations}

While performing the experiments, we noticed that the results for the zero-shot NER may be unstable. Namely, in different runs, using the same datasets and same models, we obtained different results for zero-shot NER. In a few-shot regime, the results would stabilize and would not differ much, but in zero-shot, the difference could be high. For example, for the class \textit{Drug}, in four training runs on the same data, we obtained F1 score results ranging from 0.96\% to 44.66\%.  To create reproducible results, we have fixed the \textit{seed} parameter (seed=0) and set PyTorch to use deterministic methods. The previous tables and text reported the results with the fixed seed and using deterministic methods. 

This behavior possibly occurs due to gradient descent to some of the local minima and remaining there instead of reaching the global minimum. Given that we work with text, in high dimensional vector space, our objective function is likely non-convex and complex with many local minima and maxima. Gradient descent can get stuck in a local minimum if the gradient is flat or small in that region, preventing the algorithm from exploring other parts of the search space \citep{noel2012new}. In our future work, we will explore methods to reliably reach a local minimum, such as adjusting the learning rate, adding regularization, adding momentum to the optimization algorithm, and exploring other adaptive learning rate methods.

Fine-tuning the initial model for a new class in a few-shot mode involves optimizing its parameters and configurations to enhance its performance specifically for a particular class. This targeted fine-tuning typically results in improved accuracy and precision for the designated class. However, this process often comes at the expense of other NER classes, as the model's focus shifts towards prioritizing the nuances and patterns relevant to the fine-tuned class. Consequently, while the performance for the fine-tuned class may increase, the accuracy and efficacy of identifying other NER classes may decrease due to the model's narrowed focus and potentially diminished generalization capabilities. Since our focus was on zero- and few-shot learning, evaluation of the performance of pretraining classes after fine-tuning for a new class was out of the scope of this research. However, we did some initial tests, which showed either no decrease or a slight decrease in the performance of pretraining classes (up to 3\%). However, further investigation may need to be done in the future.  

When performing entity recognition of multiple biomedical entities, traditional approaches would classify tokens in one pass. Because of our prompt engineering in the input, where the name of the class is part of the input, and transformation to binary token classification at the output, our method would need one pass per NER class, making it potentially slower to execute by increasing computational complexity.

One additional limitation of our NER approach, which classifies tokens as binary (entity/not entity), is that it is at times hard to distinguish multi-word entities from entities stated next to each other effectively. This challenge has led to the adoption of the IOB annotation schema, which differentiates between the beginning (B), inside (I), and outside (O) of entities. We chose the binary classification method because we hypothesized that learning two classes would be easier for the model than learning three classes as required by the IOB schema. However, recognizing the potential benefits of the IOB schema for handling multi-word entities, we plan to experiment with this annotation method in future work.

\section{Conclusion and future work}
\label{sec:conc}

In this paper we have presented a method for zero- and few-shot named entity recognition in the biomedical domain. However, the same method can be applied to other domains, as long as annotated data for a larger number of classes is available. The method is based on encoder-based transformer models and input data modification that factorizes regular multi-class token classification into binary classification, given the named entity class that the user would like to retrieve. The method showed state-of-the-art results for zero-shot NER and, to the best of our knowledge, it is the first method specifically shaped for zero- and few-shot NER in the biomedical domain. We have evaluated our model on 9 common biomedical classes, but the method should behave in a similar fashion for other classes.

When using our model, named entity classes do not need to be defined in advance, which is very suitable for open-set domains. Users may place any named entity class name in the first segment and the model will identify tokens in the second segment corresponding to it. This allows for zero-shot named entity recognition, where the model will recognize the entity it has not been trained on.

Our setup also enables training from any checkpoint and adding new examples with new labels without restrictions, allowing for fine-tuning of few-shot learning methodology. Providing and fine-tuning the model with several supporting examples of previously unseen classes leads to the performance improvement of the model, bringing the F1 score to 77-87\% after 100 examples. These results are often close and comparable to the BERT-based models trained on the whole dataset containing thousands of examples.

We also compare and discuss various strategies for zero-shot model validation. We have proposed validation strategies which can be summarized as (1) using unseen classes only, (2) using seen classes only, and (3) using both seen and unseen classes. 
We note that using an unseen class in validation may lead to the selection of a better model for zero-shot NER. However, since users do not know what the unseen classes will be, this method has questionable practicality.

We noticed that the base zero-shot model showed instability issues, which is likely caused by the model not being able to reach the global minimum of the optimization function. We are planning to address this issue in future work.

While our models are fine-tuned on models (BioBERT and PubMedBERT) that inherently support the recognition of multi-word entities due to their contextual understanding, optimal performance might still require additional strategies or modifications, such as enhanced post-processing rules or specialized training to handle entity boundaries effectively, either within a binary classification framework (introduce special tokens in the training data to indicate the beginning and end of entities) or by utilizing IOB annotation schema. 
We are also planning to investigate active learning approaches that may improve the results of the method even further. Our method could also help in labeling datasets while improving itself by confirming human annotation. For example, our method can be used instead of the self-correcting network in the approach proposed by \citep{ilic2022active}. Using active learning, the model can also learn to identify difficult examples and flag them for further review. 

\section*{Model and code availability}

The base model trained on all 26 classes based on PubMedBERT, with 1 in the first segment, is available at \url{https://huggingface.co/MilosKosRad/BioNER}, while the BioBERT-based model is available at \url{https://huggingface.co/ProdicusII/ZeroShotBioNER}. The code used to process data and train the models can be found at \url{https://github.com/br-ai-ns-institute/Zero-ShotNER}.

\section*{Acknowledgments}
This research was possible thanks to computational resources made available to us by the National Data Center in Kragujevac. The authors would like to thank Jelena Mitrović, Dragiša Mišković, and Branislav Kisačanin for their insightful comments. The project described in this paper was a collaboration between the Institute for Artificial Intelligence Research and Development of Serbia and Bayer A.G.

\section*{Conflicts of Interest Statement}

The project was initiated by Bayer A.G. as a research collaboration between Bayer A.G. and the Institute for Artificial Intelligence Research and Development of Serbia. Authors were paid salaries by the organization they are affiliated with. No funds, grants, or other support was received. The authors declare they have no financial interests.

\bibliographystyle{vancouver}
\bibliography{cas-refs}

\printcredits

\bio{}
\endbio

\bio{}
\endbio

\end{document}